\documentclass[preprint,authoryear]{elsarticle}
\usepackage{etoolbox}
\makeatletter
\patchcmd{\ps@pprintTitle}% <cmd>
  {Preprint submitted to}% <search>
  {}% <replace>
  {}{}% <succes><failure>
\makeatother
\usepackage{lineno,hyperref}
\modulolinenumbers[5]
\journal{Int. Seminar on Research of Information Tech. \& Intelligent Systems (ISRITI), IEEE, 2020.}

\usepackage{fancyhdr,graphicx,lastpage}% http://ctan.org/pkg/{fancyhdr,graphicx,lastpage}
\fancypagestyle{plain}{
  \fancyhf{}% Clear header/footer
  \fancyhead[L]{Resource-Aware Auto-ML}% Left header
  \fancyhead[R]{Yang, Nam, Nasr-Azadani, Tung}% Right header
  
\fancyfoot[L]{3rd ISRITI Conference, 2020.}% Left footer
\fancyfoot[R]{\thepage\  / \pageref{LastPage}}% Right footer
}
\usepackage{amsmath,amssymb,amsfonts}
\usepackage{algorithmic,algorithm}
\usepackage{graphicx}
\usepackage{textcomp}
\usepackage{xcolor}
\usepackage{subfigure}
\usepackage{nomencl}
\makenomenclature 
\usepackage[skins]{tcolorbox}
\usepackage{ctable}
\usepackage{enumitem}
\usepackage{caption,graphicx}
\usepackage{upgreek}
\usepackage{bbm}
\usepackage{mathtools}
\usepackage{hyperref}

\newcommand{\bv}[1]{\textrm{\textbf{#1}}}
\newcommand{\norm}[1]{\left\lVert#1\right\rVert}

\date{}
\begin{document}

\begin{frontmatter}
\title{Resource-Aware Pareto-Optimal Automated Machine Learning Platform}
\author[acnaddress]{Yao Yang\fnref{yao}}
\fntext[yao]{yao.a.yang@accenture.com}

\author[acnaddress]{Andrew Nam\fnref{andrew}}
\fntext[andrew]{a.nam@accenture.com}

\author[acnaddress]{Mohamad M. Nasr-Azadani\corref{corrAuthor}}
\cortext[corrAuthor]{Corresponding author. mohamad.nasr-azadani@accenture.com}
% \ead{mohamad.nasr-azadani@accenture.com}

% \author[ruiwenaddress]{Ruiwen Li\fnref{ruiwen}}
% \fntext[ruiwen]{louiseruiwenli@gmail.com}

\author[acnaddress]{Teresa Tung\fnref{teresa}}
\fntext[teresa]{teresa.tung@accenture.com}

\address[acnaddress]{Accenture Labs, 415 Mission St, Floor 34, San Francisco, CA 94105, USA}
%\address[ruiwenaddress]{Computer Science Department, Santa Clara University, Santa Clara, CA 95053, USA}

\begin{abstract}%

In this study, we introduce a novel platform Resource-Aware AutoML (RA-AutoML) which enables flexible and generalized algorithms to build machine learning models subjected to multiple objectives, as well as resource and hardware constraints. RA-AutoML intelligently conducts Hyper-Parameter Search (HPS) as well as Neural Architecture Search (NAS) to build models optimizing predefined objectives. RA-AutoML is a versatile framework that allows user to prescribe many resource/hardware constraints along with objectives demanded by the problem at hand or business requirements. At its core, RA-AutoML relies on our in-house search-engine algorithm, \textit{MOBOGA}, which combines a modified constraint-aware Bayesian Optimization and Genetic Algorithm to construct \textit{Pareto optimal} candidates. Our experiments on CIFAR-10 dataset shows very good accuracy compared to results obtained by state-of-art neural network models, while subjected to resource constraints in the form of model size. 

\end{abstract}


\end{frontmatter}

\section{Introduction}

With the significant success of deep learning models, a.k.a. Deep Neural Networks (DNN), in performing complex tasks (e.g., image recognition, natural language comprehension), various industry and government entities have begun to integrate DNNs in their, often times, complex analytical applications (for a survey, see \cite{pouyanfar2018survey}). As the complexity of the tasks along with higher accuracy requirements grew more, so did the size, model architecture complexity, and training process of new DNN models (cf. \cite{liu2017survey}). To build such models efficiently, often times due to the high cost of training a DNN, engineers have to consider many factors in the design of model architecture, training process, as well as resources to be used by the production model. Engineer tasked to build and train the best model may be faced with- but not limited to- questions such as: business goals, training budget, training data size and data type, hardware constraints, security, model inference time, and memory/power consumption.

Research has utilized advanced optimization methods augmented with sophisticated search strategies to realize automated search to return the best performing machine learning models, a.k.a. `AutoML'. Conventional AutoML algorithms developed such that their search strategy is actively guided to maximize model accuracy or similar metrics given a predefined search space for parameters of interest (for a review on AutoML algorithms, cf. \cite{zoller2019survey}). Under these circumstances, the main task of an AutoML algorithm is usually to: a) Hyper-Parameter Search (HPS): where model architecture is fixed but the value of parameters such as learning rate or batch size can define the best performing model, or b) Neural Architecture Search (NAS): where model architecture is not \textit{a priori} known and has yet to be `constructed' by an iterative search algorithm (cf. \cite{JMLR:v20:18-598}). 

More recently, training very complex DNN models, e.g. `\textit{Turing-NLG}' a DNN language model with more than 17 billion parameters introduced by Microsoft  (see \cite{turingNLG}), has become feasible. However, the enormous energy power as well as hardware requirements opened a new research in AutoML, i.e. building machine learning models considering their performance on specific hardware architectures, e.g. model pruning, model quantization, and hardware-aware acceleration of DNNs (cf. \citet{he2018amc,wang2019haq,stamoulis2020hardware,gupta2020accelerator}). 

Real-world applications may require DNN models to be agnostic to the hardware architecture, i.e. models can be trained, deployed, monitored, or re-trained on various system architectures, e.g. CPU, GPGPU, FPGA, cellular phones, cloud platforms, web-browser. This naturally imposes added limitations and constraints to building new or modification of existing models. In addition, taking into account business goals and other constraints such as budgetary limitations results in the need to make \textit{ad hoc} decisions to be made during model building or training phases.

To address the more modern requirements needed to automatically build DNN models, the existing AutoML process has yet to be expanded to include: resource/hardware constraints, multiple goals to define the `best' performing model, and robust and efficient search strategy which exhibits model candidates and the trade-offs between them and the desired goals. 
An efficient platform which can cope with the more sophisticated intricacies of modern AutoML should address issues such as: 1) how to minimize manually specifying an \textit{accurate} range for hyper-parameters in the search space by human user, 2) how to automatically integrate resource and/or hardware constraints into the search process, 3) search parallelization and scaling up model optimization given the iterative nature of building/training different models, 4) conducting complex search tasks on AutoML problems with a focus on both hyper-parameters search as well as neural architecture search subjected to the constraints concurrently, 5) appropriate tools or models to measure metrics such as memory consumption, power usage, inference time, and operational complexity, 6) transparency and flexibility in process of selecting `best' model candidates, and 7) providing the end user with accurate charts and metrics showing the trade-offs amongst different objectives and potentially best performing models, are a few example. 

To address the issues above, we introduce our generalized platform `RA-AutoML' which provides algorithms and tools to build a machine learning modeling automatically. Asking a human engineer to translate business goals along with architectural requirements into an AutoML search problem, our RA-AutoML platform asks user for any number of objectives along with any number of resource (or hardware) constraints to build their machine learning. RA-AutoML is enabled by our in-house optimization algorithm MOBOGO which utilizes a modified constraint-aware Bayesian Optimization, non-dominating sorting genetic algorithm, NSGA-II, to extract \textit{Pareto optimal} candidates, and TOPSIS algorithm to select `best' performing model, automatically. 

The current study is structured as follows. In \autoref{sec:methodology}, we present our platform RA-AutoML and its three main components along with our in-house optimization engine MOBOGA. We then proceed to present results for two sets of experiments on CIFAR-10 dataset using RA-AutoML in section \autoref{sec:experiments}. Finally, we present a summary and conclusion in \autoref{sec:summary}. Many of supporting material and algorithms are also details in \autoref{sec:appendxi}.
\section{Methodology}
\label{sec:methodology}

Imagine a machine learning engineer is tasked to design a DNN model and train it on a known training set. Additionally, business requirements demand that the final trained model should maintain very good test accuracy as well as minimal inference time. Due to hardware considerations, the final trained model has to be bounded by a set of known resource constraints, e.g. memory usage should be less than 0.5 Gb. 

\begin{figure}[htb]
\centerline{\includegraphics[width=1.0\textwidth]{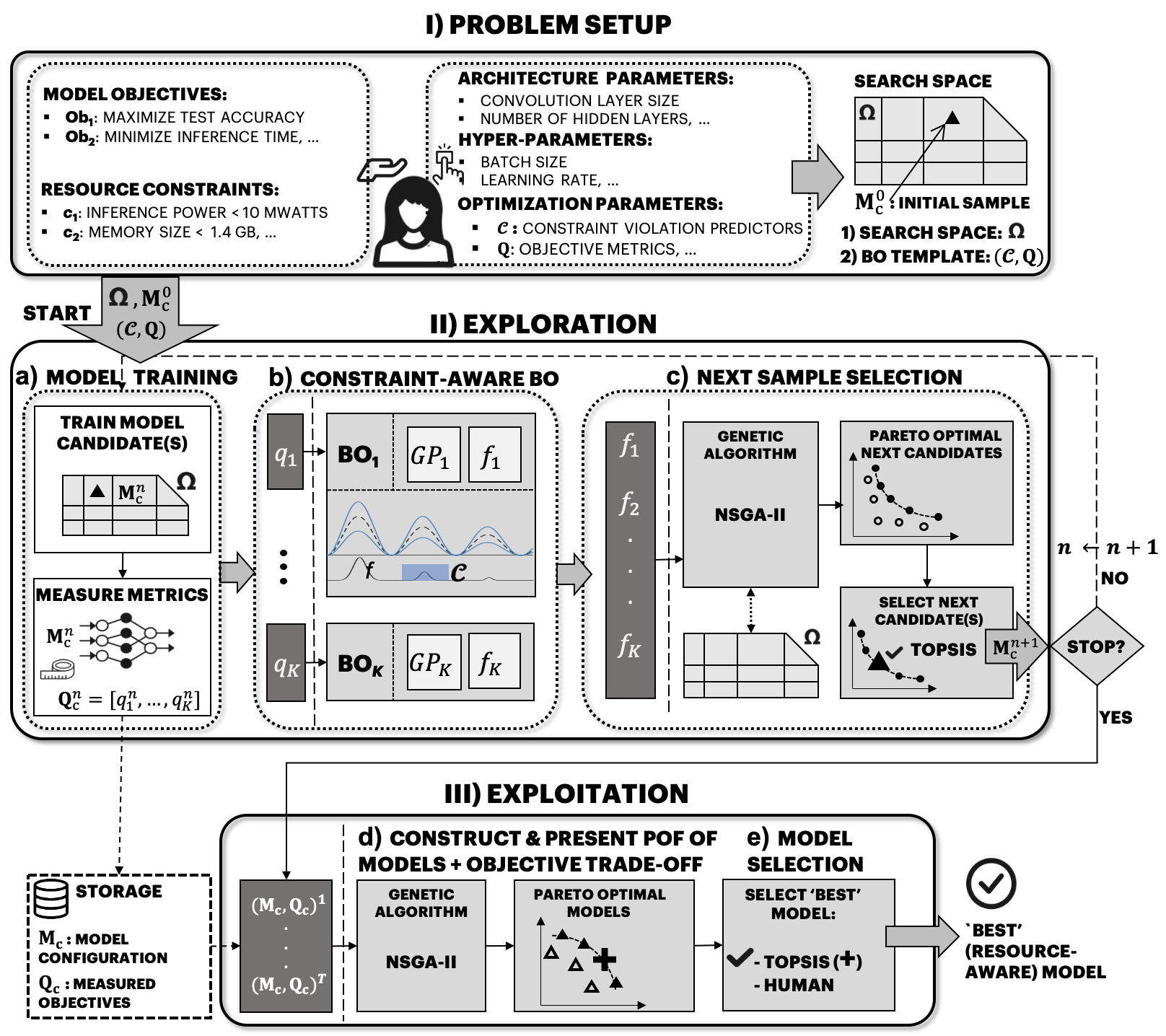}}
\caption{High level framework of our proposed framework, RA-AutoML. Our framework enables user to prescribe business and technical goals subjected to hardware or resource constraints. RA-AutoML uses a our MOBOGA, a multi-objective optimization technique to create \textit{Pareto optimal} model candidates subjected to given constraints.}
\label{fig:arch}
\end{figure}

To address examples above, our RA-AutoML platform can be utilized in a semi-automatic fashion. Towards this, RA-AutoML is equipped with three main components which can return the `best model' given problem statement. First, in `Problem setup' (Fig. \ref{fig:arch}-I), user translates business requirements and problem constraints into measurable metrics and meaningful mathematical goals that can be estimated or profiled by RA-AutoML. 
% ************************************************************
% ************************************************************

\noindent
In second step, `Exploration' (Fig. \ref{fig:arch}-II) our MOBOGA optimization engine performs intelligent search to query model candidates using BO and GA. Upon collecting enough samples, the final step is to `Exploit' (Fig. \ref{fig:arch}-III) existing samples and create set of \textit{Pareto optimal} model candidates. User is presented with clear trade-offs amongst different objectives. In the following, we provide the details for each components of our RA-AutoML framework and MOBOGA algorithm, which can address the challenges mentioned above. 

\subsection{Problem Setup}
\label{ssec:problem setup}
As the first step, engineer may be provided with various business and technical requirements, hardware choices, or even resource constraints. From the technical perspective, user translates such goals into meaningful and measurable optimization metrics for every machine learning model, e.g. model accuracy, Precision/Recall, inference time, memory consumption. 
Furthermore, any constraints ought to be formulated using proper mathematical relations, or proxy informative models. For instance, in a problem where memory should not surpass 1.5 Gb, a predictor model can be preconfigured so that it predicts whether or not a model size will violate the memory threshold. The main output of `Problem Setup' includes proper optimization cost function(s), constraint(s), and model candidate search space ($\Upomega$ in Fig. \ref{fig:arch}-I)). Other secondary choices can be also made or default values can be used, e.g. stopping criteria and initial models ($\bv{M}_c^0$). 

\subsubsection{Objectives}
\label{ssec:multiple objectives}

Given problem statement, user assigns $K$ ($\ge 1$) objectives, $\bv{O}=\{ob_1,\dotsc, ob_K\}$, ought to be optimized to return the `best' model candidates. Typically, objectives are chosen in a way that at the end of training cycle, they can be `measured', e.g. model accuracy or forward inference time. In the remainder, we denote any quantified metrics representing $K$ objectives, $\textbf{O}$, by $\textrm{\textbf{Q}} = [q_1, q_2, ..., q_K]$. Mathematically speaking, any AutoML problem can be formulated as a single ($K=1$) or multi-objective optimization task ($K > 1$) subjected to a given set of constraints with size $N_c$ ($\ge 0)$, denoted by $c_j$: 
\begin{eqnarray}
&&\textrm{\textbf{Given:}}\ \ q_1: \bv{x} \rightarrow \mathbb{R}, \dotsc,q_K: \bv{x}
\rightarrow \mathbb{R}\ \ \ (\bv{x} \in \Upomega)\, , \nonumber\\
&\textrm{Minimize}& q_1, \dotsc, q_K\, .\nonumber\\
&\textrm{subject to:}& c_1(\bv{x}), \dotsc, c_{N_c}(\bv{x})
\label{eq:multi-optimization}
\end{eqnarray}

\noindent
Since the true cost function(s) in a typical AutoML problem is not \textit{a priori} known, applying `Gradient-based' family to solve Eq. \ref{eq:multi-optimization} is not feasible in a straightforward fashion. Alternatively, one can reduce the above problem into a single objective optimization via using surrogates for a cost function. Statistical measures such as $F_\beta$-score and Mathews Correlation Coefficient (MCC), scalarization (and its variants such as inverse efficiency score) are a few techniques. (cf. for a review \cite{emmerich2018tutorial}). While these techniques can help reduce the computational cost and complexity associated with multi-objective search in problem \ref{eq:multi-optimization}, their performance and accuracy strongly depend on the choice of surrogate cost function. They can also suffer from producing sub-optimal solutions due to improper scaling, negative or fractional values. Furthermore, they remove the transparency in how model produce the `best' solution by removing a \textit{Pareto front} of `best' model candidates w.r.t various objectives. 

To address the challenges above, in our RA-AutoML framework, we employ a modified Bayesian optimization and Genetic Algorithm without the need for a surrogate cost function. BO is a derivative-free optimization technique that does not need the cost function and has numerous favorable properties (see \autoref{ssec:resource-aware-BO}). For a review of derivative-free optimization algorithms, cf. \cite{rios2013derivative}.

\subsubsection{Constraints}
\label{ssec:constraints}

Incorporating resource, hardware, and budget constraints in AutoML remains to be a challenging task. Example studies have addressed these issues with a focus on optimization of DNNs for various hardware architectures, memory limits and power constraints, cf. \citet{stamoulis2020hardware,gupta2020accelerator}, or budget constraints, cf. \cite{DBLP:journals/corr/abs-1902-00532}.

In our framework, user can impose a set of $N_c$ ($\ge 0$) constraints, namely, $\textrm{Cst}=\{c_1,c_2,...,c_{N_c}\}$ while searching for the best model(s). Let $\mathcal{C}$ denote the \textit{Boolean} function indicating whether the $i^{\textrm{th}}$ constraint, $c_i$, is satisfied for given input $\bv{x}$:
\begin{eqnarray}
\mathcal{C}(\bv{x}; c_i) = 
\begin{cases}
      \textrm{True}\ (\equiv 1);\ \ \ \ \ \ \ \ \textrm{if for $\bv{x}, \textrm{constraint}\ c_i$ is satisfied,}\\
      \textrm{False}\ (\equiv 0);\ \ \ \ \ \ \ \textrm{otherwise.}
\end{cases}  
\label{eq:constr}
\end{eqnarray}

\noindent
Here, $\bv{x}$ denotes any model candidate sampled from the search space $\Upomega$ (see \autoref{ssec:search space}). In our framework, $\mathcal{C}$ in Eq. \ref{eq:constr} is \textit{a priori} known. 
Depending on the constraint user can provide explicit rule-based or classification models (which have been trained on dataset samples for various hardware architectures), or implicit theoretical models, cf. \cite{bianco2018benchmark}). 
Our RA-AutoML framework employs constraint-aware acquisition function(s) for every Bayesian Optimization agent, $\textrm{BO}_j$, to enforce given constraints. For implementation details, see \autoref{ssec:resource-aware-BO} and \autoref{RA-EI: soft constraints}. 

\subsubsection{Search Space}
\label{ssec:search space}
Similar to traditional Auto-ML framework, user of MOBOGA defines a general search space (denoted by $\Upomega$), where model candidate, $\bv{M}_c$, can be sampled from to be queried (Fig. \ref{fig:arch}-I). In the most general form, search space may include hyper-parameters (e.g. learning rate) as well as model architecture parameters (e.g. number of neurons in a fully-connected layer, building blocks of a convolution layer). There, it is common to construct a high-dimensional space where each `point' represents a candidate model. Each axis in $\Upomega$ represents a hyper-parameter of interest. These parameters can also be of mixed types, i.e. continuous (e.g., dropout rate), discrete (e.g., number of layers), and categorical (e.g., activation function type). 
While our RA-AutoML framework can, in general, allow for a large number of hyper-parameters, to reduce computational cost and better convergence behavior, it is recommended to keep the dimension of search space dimension low by only including the most important design and/or hyper-parameters. 

\subsection{Exploration Phase}
\label{ssec:exploration}

Once the problem is formulated mathematically for {RA-AutoML}, search space $\Upomega$ has yet to be sampled  intelligently to collect enough observations, `Exploration'. In general, {AutoML} techniques have utilized various search algorithms to perform the exploration task: a) grid search and its variants, random search and batch search (cf. \citet{Bergstra:2012:RSH:2188385.2188395,park2019effect}), which exhibit unfavorable exponential scaling as the number of parameters grow, b) heuristic search methods such as naive evolution \cite{DBLP:journals/corr/RealMSSSLK17}, anneal search (cf. Hyperband \citet{DBLP:journals/corr/LiJDRT16,li2017hyperband}), Tree-structured Parzen Estimator (TPE) \cite{NIPS2011_4443}, Sequential Model-based Algorithm Configuration (SMAC) and Sequential model-based optimization (SMBO), cf. \cite{HutHooLey11-smac}, Metis Tuner \cite{li2018metis}, and more recently, c) reinforcement learning: has been adopted to perform AutoML search showing promising results, \citet{zoph2016neural,DBLP:journals/corr/SchulmanWDRK17}. 

In the exploration phase (see Fig. \ref{fig:arch}-II), we employ a Bayesian Optimization (BO) unit as well as NSGA-II, a Genetic Algorithm (GA), to traverse the model search space $\Upomega$ intelligently using feedback from the previously trained (a.k.a. queried) model candidates. Combining BO and GA enables us to address challenges with regards to non-convex cost functions as well as optimizing for more than one objective function simultaneously. In this section, we provide details of each module in Exploration Phase.

\noindent
\subsubsection{Model Training}
\label{ssec:model training lab}
At iteration $n$, model candidate(s) denoted by $\bv{M}^n_c$ is trained on existing training dataset. In addition, training agent (see Fig. \ref{fig:arch}-a) is equipped with necessary profiling tools to measure objectives $\bv{Q}^n_c$ or other relevant metrics (e.g. hardware utilization). Also, a storage system records trained models as well as observed objectives in each iteration, i.e. $(\bv{M}_c^n,\bv{Q}_c^n)$. The entire collection will later be employed in the `Exploitation phase' (see discussions in \autoref{ssec:exploitation}). 

\subsubsection{Constraint-aware Multi-Objective Bayesian Optimization}
\label{ssec:resource-aware-BO}
% Reference \cite{marculescu2018hardware} also uses hardware-aware BO... 
In our RA-AutoML framework, we adopt a modified Bayesian Optimization (BO) method to suggest next model candidate(s), $\bv{M}_c^{n+1}$, to be queried. Bayesian optimization is a powerful derivative-free optimization technique which addresses issues such as a non-convex or discontinuous shape for cost function. It also exhibits robust convergence properties along with flexibility to user through variety of choice of acquisition function (cf. \cite{rios2013derivative} for a review of derivative-free optimization algorithms and their implementations). 

In our framework, we instantiate a total of $K$ ($\ge 1$) BO agents each optimizing an objective set by user, e.g. model accuracy and inference time (see Fig. \ref{fig:arch} and \autoref{ssec:multiple objectives}). Every $\textrm{BO}_j$ ($1 \le j \le K$) uses model candidates ($\bv{x} \equiv \bv{M}_c$) and associated measured objectives (quantities, $\bv{Q}_c$) to update the surrogate function $s_j(\bv{x})$ representing cost function. In addition, acquisition function $f_j(\bv{x})$ aims to recognize next candidates in the search space for regions with highest uncertainty in surrogate function $s_j$ (cf. \cite{Frazier2018ATO}).

In RA-AutoML, $f_j(\bv{x})$ associated with $\textrm{BO}_j$, inherently satisfies resource- or other user-defined constraints. Towards this goal, we adopt a Constraint-Aware Expected Improvement (CA-EI) to construct the acquisition function (cf. \cite{gelbart2014bayesian}). For a set of constraints $c_i$, modified acquisition function is written by 
\begin{eqnarray}
f_j(\bv{x}) = \int_{-\infty}^{+\infty} \max{\{y_j^+ - y_j,0\}} \cdot p_{s,j} \left(y_j|\bv{x}\right) \cdot \prod\limits_{i=1}^{N_c} \mathcal{C}(\bv{x}; c_i) \ \textrm{d}y\, ,\ j=1,\dotsc,K\, .
\label{eq:HW-ei}
\end{eqnarray}

\noindent
Here, $y$, $y^+$, and $p_s$ denote, respectively, the value of objective function, the value of the `best' observed sample so far, and predictive marginal density of the objective function at $\bv{x}$. A binary (1 or True, 0 or False) value returned by constraint-aware indicator function (see Eq. \ref{eq:constr}) guarantees a brute force enforcement of the given constraints. In order words, the regions of model search space $\Upomega$ that violates constraints would never be explored by any $\textrm{BO}_j$ agent. User may relax strict enforcement of hard constraint and allow exploration of search space but at a lower probability, i.e. 'soft constraints'. For more details on soft-constraints, see \autoref{ssec:RA-EI verification} and \autoref{RA-EI: soft constraints}).

\subsubsection{Next Model Candidate Identification}
\label{ssec:model candidate iden}
In this step, we employ GA to recommend the future model candidate(s) $\bv{M}^{n+1}_c$ to be queried in the next iteration, $n+1$, of search space exploration. For $K \ge 2$ objectives, identifying the next set of model candidates is not straightforward since every acquisition function $f_j$ (obtained from $\textrm{OB}_j$) can recommend different model samples. To address this, given enough computational resources, user can train all candidates from $K$ acquisition functions in the next iteration. However, as the number of objectives, $K$, increases, this results in added computational cost and inefficiency in search strategy. To address this, we employ a genetic algorithm called NSGA-II (\cite{996017} to return a set of new candidates using \textit{Pareto optimality} of the most informative next candidates. NSGA-II is a non-dominated sorting-based Multi-Objective Evolutionary Algorithm (MOEA) with benefits such as reduction in computational complexity and elimination of the need to specify a sharing parameter (see \autoref{ssec:apx NSGA-II} for more details).

In our algorithm, NSGA-II utilizes acquisition functions \mbox{$\bv{F}^n = [f^n_1, f^n_2, \dotsc, f^n_K]$} at iteration $n$ (with $f_j$ given in Eq. \ref{eq:HW-ei}) to populate $J$ new potential candidate models
\begin{eqnarray}
\Lambda = \left\{(M^*_{c,1},\bv{F}_1)\, ,(M^*_{c,2},\bv{F}_2)\, ,\dotsc\, ,(M^*_{c,J},\bv{F}_{J}) \right\}\, . 
\label{eq:set_model_candidates}
\end{eqnarray}

\noindent 
Next, we proceed to construct a \textit{Pareto optimal} subset $\mathbb{P}_M$ which includes the most informative model candidates to be queried in the next iteration
\begin{eqnarray}
\mathbb{P}_M = \left\{(M^*_c,\bv{F}) \in \Lambda: \nexists\ (M^*_c,\bv{F})' \in \Lambda, \ \textrm{such that}\ \ \bv{F}' \prec \bv{F} \right\}\, .
\label{eq:pareto_front_next_model_candidates}
\end{eqnarray}

\noindent
Here, $\prec$ operator stands for \textit{Pareto domination} rule given by
\begin{eqnarray}
\textrm{Given vectors:}
\begin{cases}
\bv{V}=[v_1,v_2,...,v_K] \\
\bv{W}=[w_1,w_2,...,w_K]    
\end{cases}
\bv{V} \prec \bv{W}\ \ \Longleftrightarrow\ \ 
\begin{cases}
    \forall i \le K, v_i \le w_i\, ,\\
    \exists j \le K, v_j < w_j\, .
\end{cases}  
\label{eq:pareto_opt_definition}
\end{eqnarray}

\noindent
Finally, user can set a custom rule to pick candidate(s) from $\mathbb{P}_M$ or let MOBOGA automatically pick the most informative sample configuration $\bv{M}^{n+1}_c$ from Pareto set $\mathbb{P}_M$ (Eq. \ref{eq:pareto_front_next_model_candidates}) using the TOPSIS algorithm (see \autoref{ssec:apx-topsis}).

\subsubsection{Stop Exploration}\label{ssec:stop exploration}
In our platform, user can either set explicit rules, e.g. maximum number of iterations, or employ an automated and intelligent metric to stop the exploration. In doing so, at every iteration in the Exploration step, the minimum `distance' between the next model candidate $\bv{M}^{n+1}_c$ and existing (queried) models is calculated and compared to a predefined threshold $\delta$
\begin{eqnarray}
d = \min{\{\norm{\bv{M}_c^{n+1} - \bv{M}^i_c}\}}\, ,&&\ \  i=0,1,\dotsc,n\, ,\nonumber\\
\textrm{Stop exploring if:}\ \ \ &d& \le \delta\, .
\label{eq:stopcriteria}
\end{eqnarray}

\noindent
If stoppage criterion is satisfied, `Exploration' is terminated and MOBOGA proceeds to the `Exploitation' step (see Fig. \ref{fig:arch}-III). Otherwise, the new model candidate $\bv{M}_c^{n+1}$ will be queried by the `Exploration unit'. We further remark that one can also include training budget or other metrics as other `constraints' (cf. \cite{DBLP:journals/corr/abs-1807-01774}) in the search process to impose budgetary constraints.

\subsection{Exploitation and Pareto Optimal Model Candidates}
\label{ssec:exploitation}
Once stoppage criteria in the Exploration step satisfied, no more model candidates is queried. Instead, `Exploitation' unit retrieves stored data including model candidates $\bv{M}_c$ and their measured objectives $\bv{Q}_c$ into a new set
\begin{eqnarray}
\Gamma = \left\{(\bv{M}^1_c,\bv{Q}^1_c)\, ,(\bv{M}^2_c,\bv{Q}^1_c)\, ,\cdot \cdot \cdot\, ,(\bv{M}^T_c,\bv{Q}_c^T) \right\}\, . 
\label{eq:set_model_candidates}
\end{eqnarray}

\noindent 
Here, $T$ denotes the total number of queried models in the Exploration step. 
As discussed in \autoref{ssec:multiple objectives}, generally speaking, one model candidate may not optimize every required objective concurrently. To address this, we construct a set of POF based on measured objectives
\begin{eqnarray}
\mathbb{P}_O = \left\{(\bv{M},\bv{Q}) \in \Gamma: \nexists\ (\bv{M},\bv{Q})' \in \Gamma,\ \  \textrm{such that}\ \ \bv{Q}' \prec \bv{Q} \right\}\, ,
\label{eq:pareto_front_obj}
\end{eqnarray}

\noindent
using NSGA-II algorithm (\cite{996017}). In RA-AutoML, we retain flexibility and transparency by constructing a set of potentially `best' candidates through extracting a POF of model candidates. By definition, objectives set by user can demonstrate competing behavior which, in turn, makes it impossible to optimize every objective simultaneously, e.g. the most accurate model may not be the fastest model. Thus, having a transparent POF of models can present the trade-off between various objectives and model candidates to the user of platform.

\subsubsection{Choosing the \textit{`BEST'} Model}
\label{ssec:topsis}
After a POF of optimal models constructed (see Eq. \ref{eq:pareto_front_obj}), we proceed to recommend the `best' model to the human user chosen automatically. In our framework, we utilize Technique for Order of Preference by Similarity to Ideal Solution (TOPSIS), cf. \cite{hwang1981multiple}, to identify the best candidate(s). This process is detailed in \autoref{ssec:apx-topsis}. Alternatively, user can pick the `best' model by inspecting the constructed POF along with considering other business goals. For instance, for a given problem, it maybe acceptable to compromise on model accuracy by 1\% if the consumed memory is reduced by 5\%. 

\subsection{RA-AutoML: Overall workflow}
\label{ssec:overall workflow}
\begin{figure} [h!]
    \centering
    \includegraphics[width=1.0\textwidth]{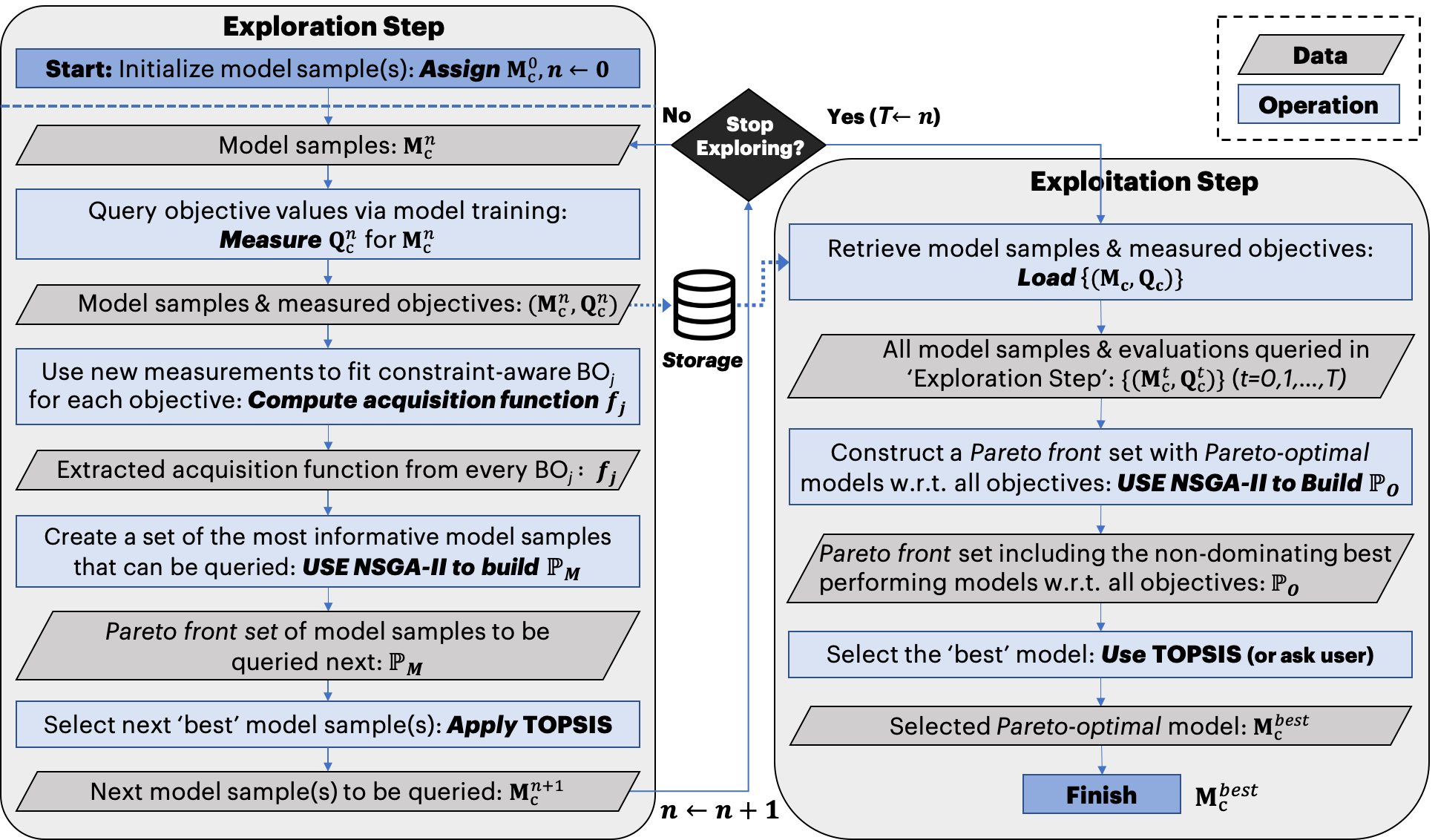}
    \caption{Overall workflow demonstrating how our RA-AutoML framework is implemented. For details see \autoref{sec:methodology}.}
    \label{fig:overall_workflow}
\end{figure}

To recap, our RA-AutoML framework consists of three main components: 
\begin{enumerate}[nolistsep]
    \item \textbf{Problem Setup}: User translates business and technical requirements into proper mathematical objectives (or point to a range of available metrics, e.g. accuracy, loss, cross-entropy loss, top-1 accuracy, top-5 accuracy, etc.), resource and hardware considerations, and RA-AutoML model search space (Fig. \ref{fig:arch}-I). 
    \item \textbf{Exploration Step}: MOBOGA optimization engine measures the multiple objectives for model candidates sampled from the search space $\Upomega$, intelligently (Fig. \ref{fig:arch}-II). 
    \item \textbf{Exploitation Step}: MOBOGA constructs a POF consisting of \textit{Pareto optimal} model candidates. User can pick `best model from TOPSIS algorithm or inspect the trade-offs amongst various candidates to choose the appropriate `best' model concerning their needs (Fig. \ref{fig:arch}-III). 
\end{enumerate}
A step by step workflow of RA-AutoML including 'Exploration' and `Exploitation' steps is illustrated in Fig. \ref{fig:overall_workflow}.

\section{Experiments and Results}
\label{sec:experiments}
In this section, we present results of two experiments obtained using our RA-AutoML platform. 
The main task is to perform image classification using CIFAR-10 dataset (\cite{cifar10}) via a DNN. First experiment is focused on HPS in the context of transfer learning while the second setup builds a DNN via NAS and HPS search, simultaneously. In both experiments, we set three objectives ($K=3$) as following: 
\begin{eqnarray}
    \bv{O} = 
\begin{cases}
    ob_1: \textrm{Minimize model `Top-5' validation error,} \\
    ob_2: \textrm{Minimize model memory consumption,}\\
    ob_3: \textrm{Minimize model forward inference.} 
\end{cases}
\label{eq:experiment_objectives}
\end{eqnarray}

\noindent
To minimize forward inference in $ob_3$, we use floating-point operations (FLOPs) as the proxy metric following models in \cite{bianco2018benchmark}). In both experiments, we apply one resource constraint, $c_1$: trained model should not consume memory more than $80\ \textrm{Mb}$ (experiment 1) and $1,400\ \textrm{Mb}$ (experiment 2), respectively.

\subsection{Experiment 1: Hyper-parameter Search in Transfer-learning }
\label{ssec:HPS-results}
In this experiment, user is asked to use RA-AutoML to perform transfer learning using existing pre-trained DNN models: VGG-16 or VGG-19 (\cite{simonyan2014very}). 
% ************************************************************
% Table
% ************************************************************
\begin{table} [ht]
\centering
\caption{Hyper-parameters returned by RA-AutoML for experiment 1 in \autoref{ssec:HPS-results}) using three objectives and one resource constraint. Best model candidate is chosen from a POF of model candidates using TOPSIS algorithm.}
\begin{tabular}{lll||l}
\hline \hline
Parameter & Parameter & Range & Best Parameter\\
Name &  Type &  &  Value\\
%heading
\hline
dropout rate & continuous & $[0,1]$ & 0.006 \\
added-layer count & discrete & (1,2,3,4,5) & 1 \\
batch size & discrete & (32,64,128) & 64 \\
activation function & categorical & \{ReLU, Tanh\} &  Tanh \\
base model & categorical & \{VGG-16, VGG-19\} &  VGG-16\\
\hline
\end{tabular}
\label{table:results-HPS} 
\end{table} 
% ************************************************************
% End Table
% ************************************************************

\noindent
To do so, we add an unknown number of fully-connected layers with equal number of neurons at the end of the base model while freezing already trained parameters. Table \ref{table:results-HPS} provides hyper-parameters used to define RA-AutoML search space $\Upomega$. Configured by user, hyper-parameters are of mixed types: categorical, discrete, and continuous.

\subsection{Generalized Search: Hyper-parameter and Neural Architecture Search} 
\label{ssec:GS}
Performing AutoML in the context of model architecture design has gained a lot of interest recently, cf. \citet{JMLR:v20:18-598,DBLP:journals/corr/TargAL16}). 
% ************************************************************
% Table
% ************************************************************
\begin{table} [ht]
\centering
\caption{Comparison of Top-5 error from existing ResNet networks and the `best' models returned by platform RA-AutoML. For problem setup, see \autoref{sec:experiments}.}
\begin{tabular}{lllll}
\hline \hline
DNN & Study & Top-5 & Memory & FLOPs\\
Network & & Error (\%) & (Mb) & \\
%heading
\hline
     $\textrm{ResNet18}$ & \cite{bianco2018benchmark} & 10.76 & N/A & N/A \\
     $\textrm{ResNet34}$ & \cite{bianco2018benchmark} & 8.74 & N/A & N/A\\ 
\hline     
     Experiment 1  & \textbf{RA-AutoML} & 7.33 &  57.15 & 8.302 G \\
     Experiment 2  & \textbf{RA-AutoML} & 8.73 & 1.155 & 4.034 M \\
\hline
\end{tabular}
\label{table:results} 
\end{table} 
% ************************************************************
% End Table
% ************************************************************

\noindent 
Integrating Neural Architecture Search (NAS) into AutoML often times adds to the complexity of search strategy for best model architecture. More recently, utilization of basic `building blocks' of DNNs to perform AutoML and NAS has demonstrated promising results via a higher efficiency, and a significant reduction in search complexity (\cite{DBLP:journals/corr/abs-1801-08577}). Towards this, an AutoML algorithm replaces searching for the best design parameters in a layer-by-layer by configuring and connecting predefined `building blocks' (SMBO \cite{HutHooLey11-smac}), e.g. a ResNet block (cf. \cite{he2016deep}).
\begin{table} [ht]
\centering
\caption{Best parameters returned by RA-AutoML for experiment 2 in \autoref{ssec:GS}) using three objectives and one constraint. Best model candidate is chosen from a POF of model candidates using TOPSIS algorithm.}
\begin{tabular}{lll||l}
\hline \hline
Parameter& Parameter& Range/Value & Best Parameter \\
Name & Type & & Value \\
%heading
\hline
activation function & categorical & \{ReLU, Leaky ReLU\} & Leaky ReLU \\
learning rate & continuous & (0,1] & 0.117\\
block type & categorical & \{Basic, Modified\} & Modified\\
block depth & categorical & \{[2,2,2,2], [3,4,6,3], & [3,4,6,3] \\
& & [3,4,23,3], [3,8,36,3]\} & \\
\hline
\end{tabular}
\label{table:results-NAS} 
\end{table} 
We adopt a similar approach to setup experiment 2 whose goal is to build a DNN model from `scratch' using ResNet blocks. We narrow the architecture search to two types of ResNet blocks (see Table 1 in \cite{he2016deep}) along with other hyper-parameters. 
It is worth mentioning that the results in this experiment could have included more variety of 'building blocks', i.e. one can add other types of existing blocks, e.g. Inception block, design their blocks based on their application and the complexity of NAS task. 
In summary, in experiment 2, search space is expanded to enable two tasks in our platform, i.e. HPS and NAS. Table \ref{table:results-NAS} summarizes search space configuration and the parameters obtained for the `best' model returned by our RA-AutoML using TOPSIS.

\subsection{Pareto-Optimal Model Candidates}
% ************************************************************
% ************************************************************
\begin{figure}[htb]
\centerline{\includegraphics[width=0.75\textwidth]{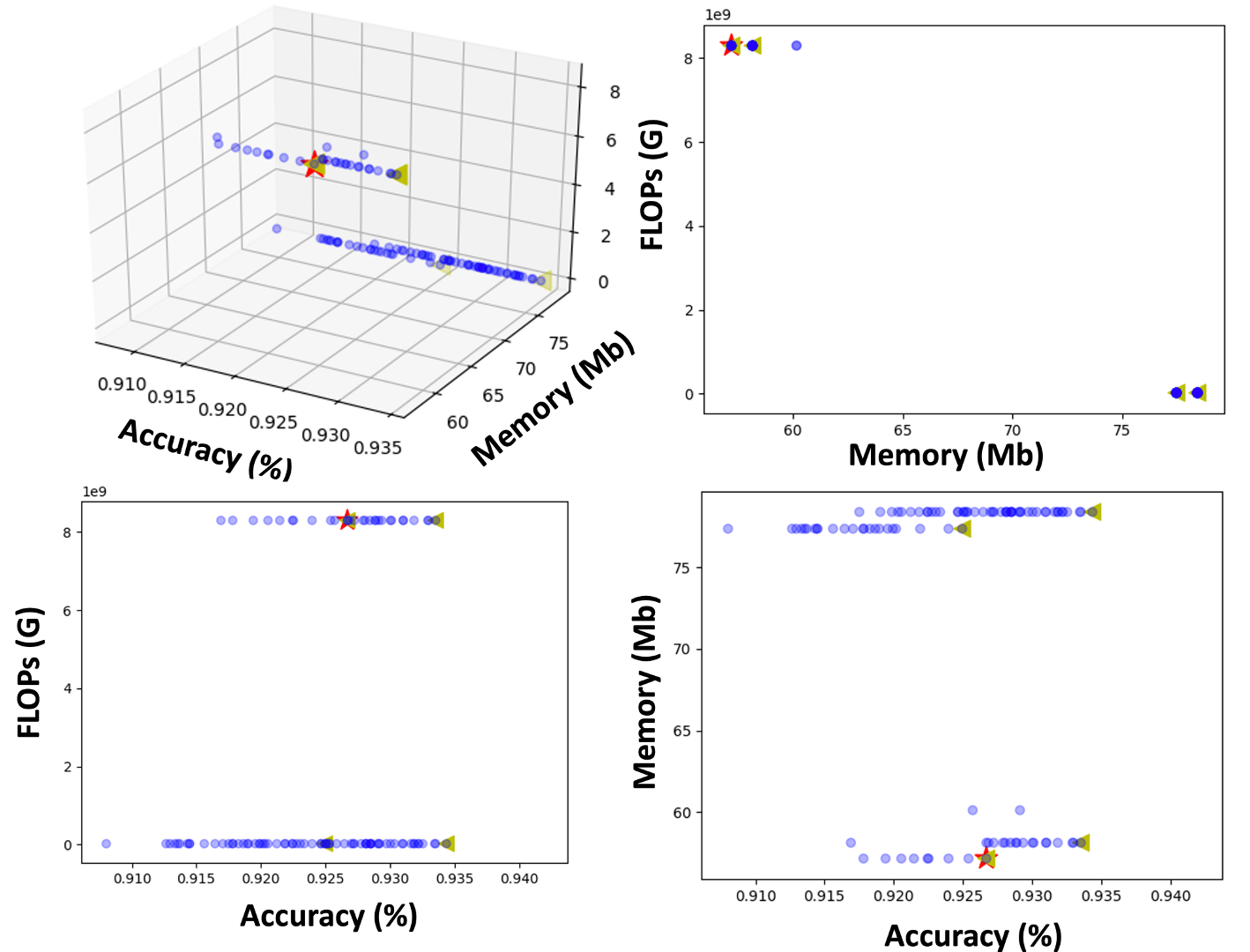}}
\caption{\textit{Pareto optimal} constructed from results for experiment 1 (described in \autoref{ssec:HPS-results}) our platform, RA-AutoML. Three-dimensional view shows all model candidates queried in the Exploration step. Yellow triangle ($\blacktriangleleft$), blue sphere ($\bullet$), and red star ($\bigstar$) correspond to, POF model candidates, trained models in the exploration phase, and the `best' model suggested by TOPSIS algorithm, respectively.}
\label{fig:pareto-front-hps}
\end{figure}

\noindent
In real world applications of DNNs where multiple goals and resources are to be considered to choose a model for deployment, it is imperative to provide a human user with transparent and flexible means so that she/he can, in turn, inspect the trade-offs amongst different goals for candidate optimal models. In our RA-AutoML platform, we present a POF consisting of \textit{Pareto optimal} model candidates retrieved in the Exploitation step along with the trade-off in measured objectives for every model candidate (see \autoref{ssec:exploitation}). 

\begin{figure}[htb]
\centerline{\includegraphics[width=0.7\textwidth]{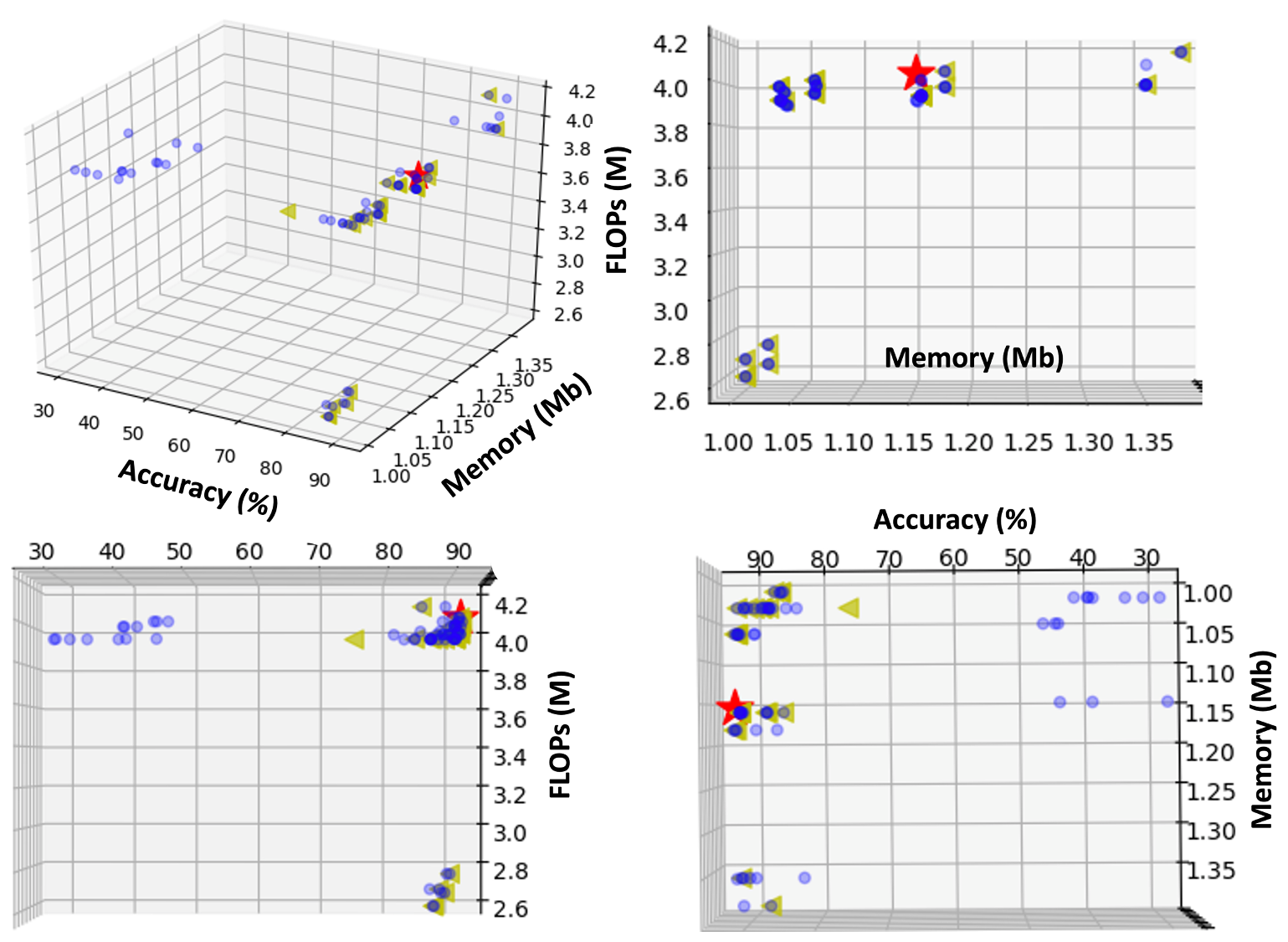}}
\caption{\textit{Pareto optimal} constructed from results by MOBOGA in Exploitation step. 3D view shows all model candidates suggested by BOs in the Exploration step. Yellow triangle ($\blacktriangleleft$), blue sphere ($\bullet$), and red star ($\bigstar$) correspond to, \textit{Parteo front} model candidates, trained models in the exploration phase, and the `best' model suggested by TOPSIS algorithm, respectively. For problem setup, see \autoref{ssec:GS}.}
\label{fig:pareto-front-nas}
\end{figure}

\noindent
Figures \ref{fig:pareto-front-nas} and \ref{fig:pareto-front-hps} present POFs created by our RA-AutoML platform for experiments 1 and 2. It is insightful to better understand how Exploration step in our platform honors the the hard constraints imposed, i.e. memory constraint when running that specific configuration while maximizing the objectives three objectives (see Eq. \ref{eq:experiment_objectives}).

\section{Summary and Conclusions}
\label{sec:summary}

Current work we have introduced RA-AutoML, a flexible, transparent and robust framework to perform automated machine learning with a focus on addressing more recent real world challenges such as multiple goals, resource constraints, and hardware-aware model optimization. Unlike traditional AutoML algorithms, our framework provide problem setup, exploration and exploitation to build an end-to-end model building (see Figs. \ref{fig:arch} and \ref{fig:overall_workflow}). In particular, our in-house optimization engine MOBOGA can combine hyper-parameter search as well as neural architecture search subjected to any arbitrary number of constraints and optimization objectives. 

We have employed our solution to build and train image detection classifiers on CIFAR-10 dataset with three objectives: Top-5 accuracy, memory consumption, and inference complexity, subjected to one resource constraint. Our algorithm creates a POF of model candidates based on exploration-exploitation modules of MOBOGA presented to human along with suggestion of the `best' model using TOPSIS algorithm. Human user can, in turn, inspect the trade-off between different \textit{Pareto optimal} models and either pick the best model manually or accept the automatically returned `best' model by RA-AutoML. 

In the next steps, it is versatile to enable online and automated hardware profiling systems hosting RA-AutoML. This way, RA-AutoML could have access to a knowledge base and learns the model configuration, hardware profile, and corresponding hardware metrics (e.g. run-time, memory use, FLOPs). This information ultimately can be used to apply CA-EI acquisition function (see Eq. \ref{eq:HW-ei}) of the Bayesian Optimization in a fully-automated manner.

\section*{Acknowledgment}
We thank Mr. Zaid Tashman Principal R\&D Scientist at Accenture Labs for sharing his insightful comments on constrained Bayesian Optimization and Ruiwen Li, our former summer intern who helped with the initial explorations.
% Also, we would like to acknowledge the Universe for being created and getting us thus far.

\section*{}
\vskip 0.2in
% \bibliography{References}
% arXiv only accepts references from *.bbl file 
% copied in a tex file

\bibliographystyle{plainnat}

\newpage
\appendix
\section{\\Experiments}
\label{sec:appendxi}

\subsection{CA-EI: Enforcing Soft-constraints}
\label{RA-EI: soft constraints}
In addition to hard-constraints, our algorithm is capable of integrating soft-constraints where a constraint-violating region in the search space alters the BO's acquisition function such that this region is can still be explored by the BO.
Examples of scenarios for soft-constraints is where user is not strict about their resource-constraint bounds, or not even very sure of precise limit values. To implement soft-constraints in RA-AutoML, first we define a new function $\mathcal{S}$ which penalizes the acquisition function by a factor $\beta$ in the regions violating a soft-constraint rule
\begin{eqnarray}
\mathcal{S} (\mathcal{C}(\bv{x}; c_i)) = 
\begin{cases}
      \textrm{1}\ \ \ \ \ \ \ \textrm{if for $\bv{x}\, ,c_i$ is satisfied}\, ,\\
      \textrm{\begin{math} \beta \end{math}}\ \ \ \ \ \ \ \textrm{otherwise, where $\beta \in [0,1)$}\, .
\end{cases}  
\label{eq:soft-constr}
\end{eqnarray}

\noindent
Factor $\beta$ can be defined as any arbitrary function can return a constant value or vary depending on the `severity' of violation, e.g. distance to the region bounds. Examples of such functions include inverse, polynomial, or exponential decay. Thus, CA-EI acquisition function (in Eq. \ref{eq:HW-ei}) is updated in the following way:
\begin{eqnarray}
f_j(\bv{x}) = \int_{-\infty}^{+\infty} \max{\{y_j^+ - y_j,0\}} \cdot p_{s,j} \left(y_j|\bv{x}\right) \cdot \prod\limits_{i=1}^{N_c} \mathcal{S} (\mathcal{C}(\bv{x}; c_i)) \ \textrm{d}y\, ,\ j=1,...,K\, .
\label{eq:HW-ei-soft}
\end{eqnarray}

\noindent 
We remark that for $\beta=0$, a soft-constraint automatically acts as a hard-constraint.

\subsection{Example of Constraint-Aware Bayesian Optimization}
\label{ssec:RA-EI verification}

\begin{figure}[tb]%[htbp]
\centering
\includegraphics[width=0.75\textwidth]{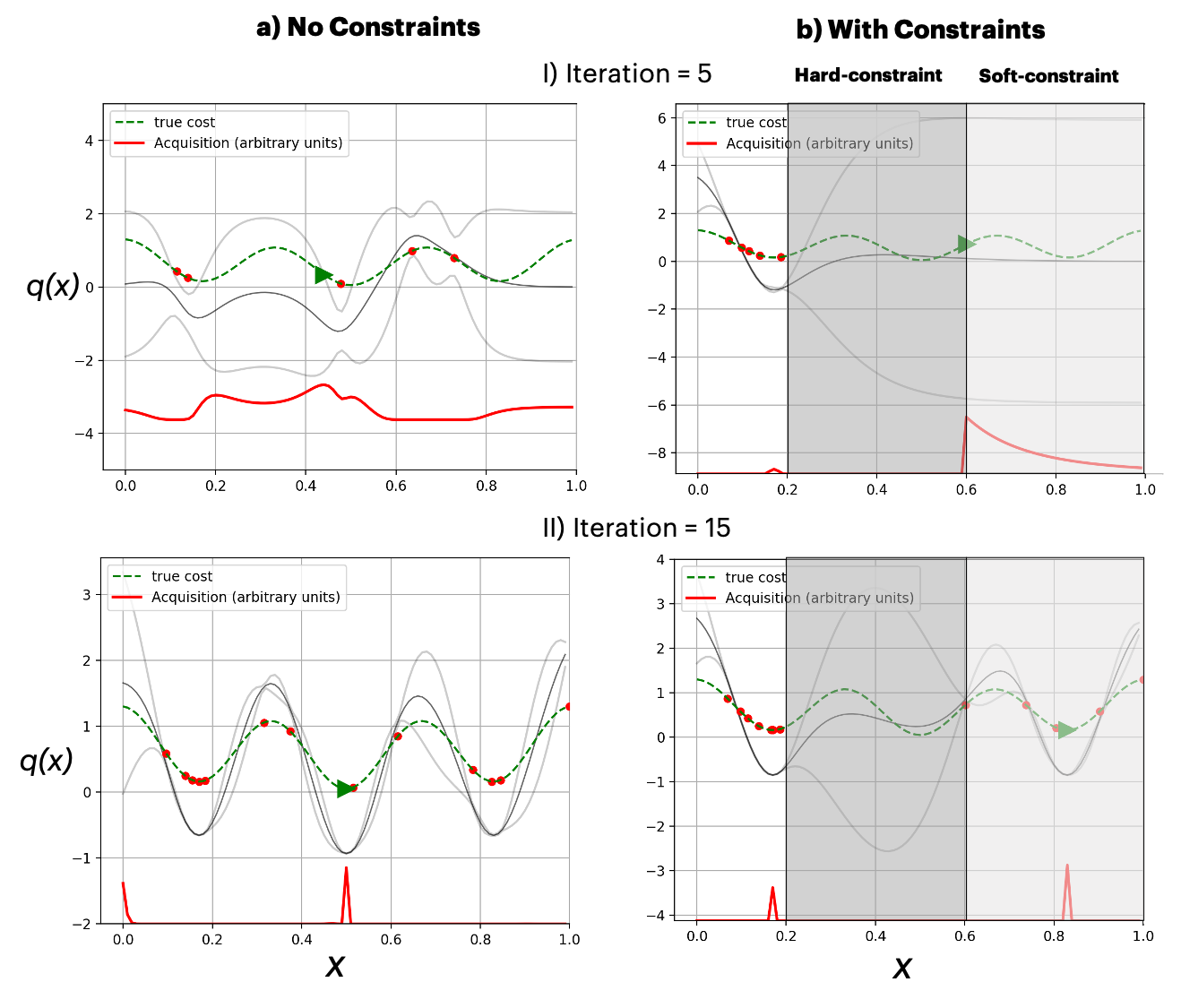}
\caption{Comparison of a constraint-free (Left) against constraint-aware Bayesian Optimization of a known objective function (Eq. \ref{eq:acq_verif_func}). The dotted line, red line, black line, black diamonds, and blue diamond, respectively, show the true cost function, BO's acquisition function, mean estimate of objective function using BO's surrogate function, and next point suggested for next iteration. Initial observation point is at $x^0=0.1$. See \autoref{ssec:RA-EI verification} for details.}
\label{fig:CA-EI-example}
\end{figure}

\noindent
To demonstrate how our modification of the acquisition function in BO works, we apply the modified Bayesian Optimization to find the minimum $x$ of a known cost function with several local minima:
\begin{eqnarray}
q(x) = 1.1 + (x-0.5)^2+\frac{1}{2}\sin{\left(6\pi x+\frac{\pi}{2}\right)}\ \ \ \ \ \ \ \ x \ge 0\, .\label{eq:acq_verif_func}
\end{eqnarray}

\noindent
We subject the optimization task to both hard-constraints in region $0.2 \le x \le 0.6$, and soft-constraints in region $0.6 < x$. To enforce the soft-constraints, penalizing factor $\beta$ (in Eq. \ref{eq:HW-ei-soft}) is defined by $\beta = 1/(x-0.6)^4$.

Figure \ref{fig:CA-EI-example} compares the solution of a constraint-free BO (left) with constraint-aware BO solution. Our constraint-aware modified Bayesian optimization has the ability to choose data points that meet the constraint requirements while honoring different types of constraints, i.e. at $15^{\textrm{th}}$ iteration the constrained BO does not explore the regions with hard-constraints, but is able to query a few samples in the regions with soft-constraints. 
\subsection{MOBOGA Verification}
\label{ssec:apx MOBOGA verification}
\begin{figure}[htbp]
\centering
\includegraphics[width=0.90\textwidth]{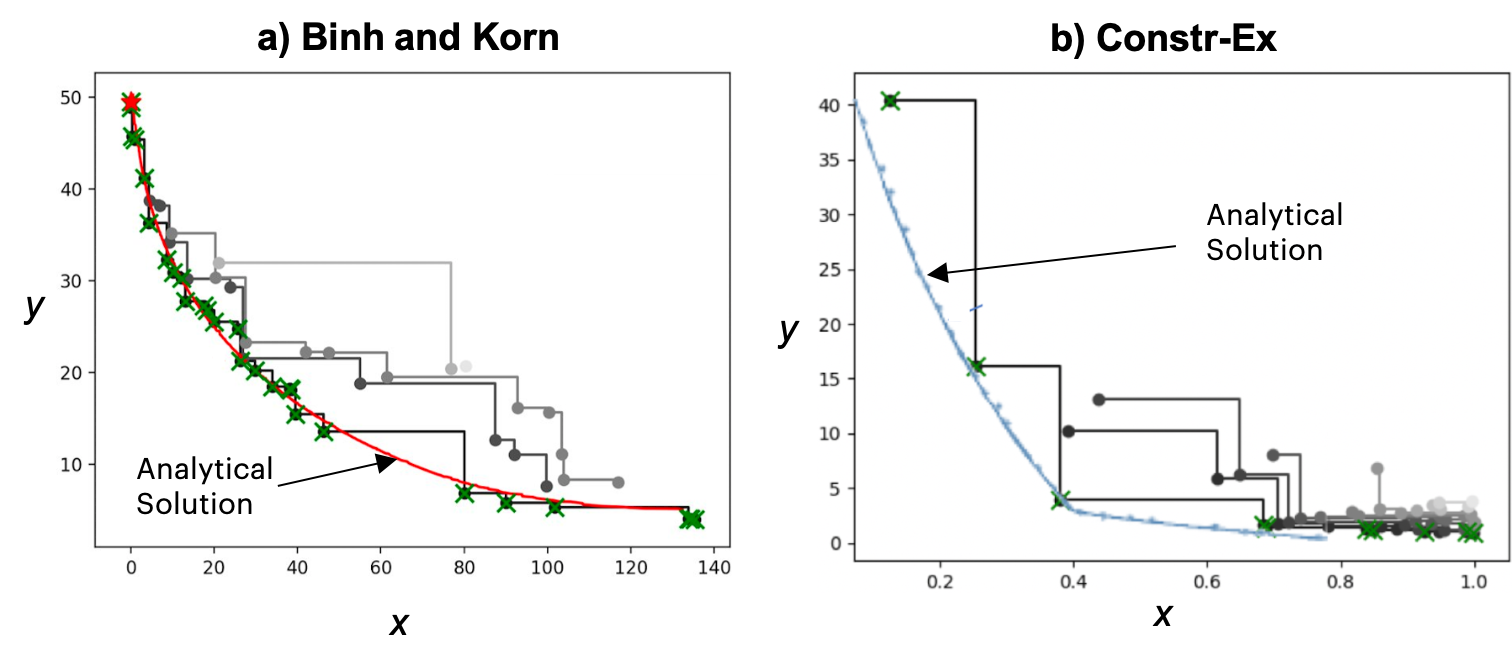}
\caption{Comparison of \textit{Pareto front} obtained by MOBOGA against analytical solution given for two multi-objective minimization subjected to known constraints (see Eqs. \ref{eq:KornBinh} and \ref{eq:constr-ex}). Green `$\times$' symbols show \textit{Pareto optimal} points after 50 iterations conducted in the Exploration step (see Fig. \ref{fig:arch}). Results are in good agreement with analytical solution.}  
\label{fig:constr-exKornVerification}
\end{figure}

In this section, we test our MOBOGA optimization to solve two known constrained and multi-objective cost functions. We further create and compare the extracted POFs using NSGA-II algorithm employed in our RA-AutoML framework. We choose Binh-Korn function (see \cite{binh1997mobes}) 
\begin{equation}
\min_{(x,y)\in \mathbb{R}^2}
\left[
    \begin{array}{l}
      q_1\left(x,y\right) = 4x^{2} + 4y^{2} \\
      q_2\left(x,y\right) = \left(x - 5\right)^{2} + \left(y - 5\right)^{2} \\
    \end{array}
\right]\, ,
\ \ \ \ \text{subject to:}
\begin{cases}
      c_1: (x - 5)^2 + y^2 \leq 25 \\
      c_2: (x - 8)^2 + (y + 3)^2 \geq 7.7 \\
\end{cases}\, ,
\label{eq:KornBinh}
\end{equation}

\noindent
and Constr-Ex with two objective functions
\begin{equation}
\min_{(x,y)\in \mathbb{R}^2}
\left[
    \begin{array}{l}
      q_1\left(x,y\right) = x \\
      q_2\left(x,y\right) = \frac{1 + y}{x} \\
    \end{array}
\right]\, ,
\ \ \ \ \text{subject to:}
\begin{cases}
      c_1:y + 9x \geq 6 \\
      c_2:-y + 9x \geq 1 \\
\end{cases}\, ,
\label{eq:constr-ex}
\end{equation}

\noindent
to conduct constraint-aware multi-objective minimization. Figure \ref{fig:constr-exKornVerification} shows results obtained by MOBOGA after 50 iterations with analytical solution. These results demonstrate very good agreement on reproducing the POFs shown by the solid lines.

\section{\\Algorithms}
\label{ssec:apx algorithms}
In this section, we summarize existing algorithms utilized in our framework.
% \subsection{Bayesian Optimization}
% \label{ssec:apx-BO}
% \begin{algorithm}\captionsetup{labelfont={sc,bf}, labelsep=newline}
% \begin{algorithmic}
% \scriptsize
% \STATE Place a Gaussian process prior on $f$ 
% \STATE Observe $f$ at $n_0$ points according to an initial space-filling experimental design. Set $n$=$n_0$. 
% \WHILE{n $\le$ N } 
    %\STATE Update the posterior probability distribution on $f$ using all available data
%    \STATE Let $x_n$ be a maximizer of the acquisition function over $x$, where the acquisition function is computed using the current posterior function
    %\STATE Observe $y_n = f(x_n)$
%    \STATE Increment $n$
%\ENDWHILE
%\STATE Return a solution: either the point evaluated with the largest $f(x)$, or the point with the largest posterior mean.
%\end{algorithmic}
%\caption{Bayesian Algorithm}
%\label{alg:BO}
%\end{algorithm}

\subsection{NSGA-II Genetic Algorithm}
\label{ssec:apx NSGA-II}
NSGA-II is a non-dominated sorting-based Multi-Objective Evolutionary Algorithm (MOEA), cf. \cite{horn1994niched}. Evolutionary Algorithms (EA)- inspired by biological evolution processes, i.e. reproduction, mutation, recombination, and selection- aim to find the `best' candidate(s) in a `population' of potential candidates. EAs can be used in multi-objective optimization problems with non-convex cost functions, cf. \cite{horn1994niched}. In doing so, EA generates a new set of candidate points, evaluate their `cost or value' given respective fitness functions, and repeat this process until termination. Here, the best, aka `fittest' points are the ones that survive in each 'iteration'. 
\begin{figure}[h!]
    \centering
    \includegraphics[width=0.55\textwidth]{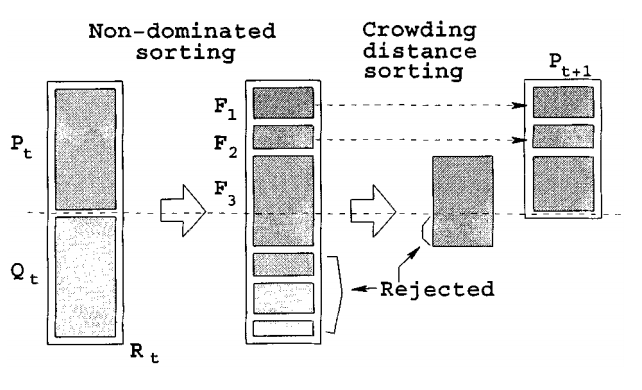}
    \caption{Schematic illustration of the Genetic Algorithm NSGA-II developed by \cite{996017}.}
    \label{fig:nsgaii}
\end{figure}
Numerous MOEA algorithms use non-dominated sorting with a few drawbacks: (1) their $\mathcal{O}\left(K \cdot N^3\right)$ computational complexity (where $K$ is the number of objectives and $N$ is the population size); (2) their non-elitism approach; and (3) the need to specify a sharing parameter (cf. \cite{996017}). 
\begin{algorithm}[tb]
\captionsetup{labelfont={sc,bf}, labelsep=newline}
\begin{algorithmic}
\scriptsize
\FOR{each $p\in P$}  
    \STATE {$S_p=0$}
    \STATE {$n_p=0$}
    \FOR{each $q\in P$}
        \IF{$p\prec q$ //\textit{if p dominates q}}  
        \STATE $S_p=S_p\cup{q}$ //Add q to the set of solutions dominated by p 
        \ELSIF{$q\prec p$}
            \STATE $n_p=n_p + 1$ // Increment the domination counter of p
            \ENDIF
    \IF{$n_p=0$  //p belongs to the first front}
        \STATE{$p_{rank}=1$}
        \STATE {$F_1=F_1\cup {p}$}
    \ENDIF
    \ENDFOR
\ENDFOR
    \STATE {i=1 //initialize the front counter}
    \WHILE{$F_i\neq 0$}
    \STATE{Q=0 // Used to store the members of the next front}
    \FOR{each $p\in F_i$}
        \FOR{each $q\in S_p$}
        \STATE {$n_q=n_q-1$}
        \IF{$n_q=0$ // q belongs to the next front}
        \STATE{$q_{rank}=i+1$}
        \STATE {$Q=Q\cup{q}$}
        \ENDIF
        \ENDFOR
    \ENDFOR
    \STATE{i=i+1}
    \STATE{$F_i=Q$}
    \ENDWHILE
\end{algorithmic}
\caption{Non-dominated Sorting in NSGA-II Genetic Algorithm}
\label{alg:NSGA}
\end{algorithm}

In RA-AutoML, we employ NSGA-II (\cite{996017}) which can address problems mentioned above. Initially, a random parent population $P$ is created, the population is sorted based on non-domination. Each solution is assigned a fitness value equal to its non-domination level. Then, it applies a binary tournament selection, followed by recombination, and mutation operators to create an offspring population $Q_0$ of size $N$ in order to minimize the fitness. Algorithm \ref{alg:NSGA} demonstrates how NSGA-II performs the non-dominant sorting. After non-dominance sorting is performed, NSGA-II algorithm can be implemented in the following manner (see Algorithm \ref{alg:NSGA2}. The NSGA-II workflow diagram is illustrated in Fig. \ref{fig:nsgaii}. The new population $P_{t+1}$ of size N is now used for selection, crossover, and mutation to create a new population $Q_{t+1}$ of size N. 

\begin{algorithm}[tb]
\captionsetup{labelfont={sc,bf}, labelsep=newline}
\begin{algorithmic}
\scriptsize
\STATE \textbf{Step 1}: 
A combined population $R_t=P_t \cup Q_t$ is formed. The population $R_t$ is of size 2N. 
\STATE \textbf{Step 2}: The population $R_t$ is sorted according to non-domination, since all previous and current population members are included in $R_t$, we can ensure elitism.

\STATE \textbf{Step 3}: If the size of $F_1$ is smaller than N, we choose all members of set $F_1$ for the new population $P_{t+1}$. The remaining members of the population $P_{t+1}$ are chosen from subsequent non-dominated fronts in the order of the ranking. Thus, solutions from the set $F_2$ are chosen next, followed by solutions from set $F_3$, and so on. Continue the procedure until no more sets can be accommodated. 
\end{algorithmic}
\caption{NSAG-II Algorithm}
\label{alg:NSGA2}
\end{algorithm}

\subsection{TOPSIS Algorithm}
\label{ssec:apx-topsis}
After we construct a POF consisting of \textit{Pareto optimal} points, RA-AutoML can proceed to recommend the `best' data point, automatically. In our algorithm, we utilize Technique for Order of Preference by Similarity to Ideal Solution (TOPSIS), see \cite{hwang1981multiple}, to extract the best candidate(s). The process is carried out as follows:

\begin{algorithm}[tb]
\captionsetup{labelfont={sc,bf}, labelsep=newline}
\begin{algorithmic}
\scriptsize
\STATE \textbf{Step 1}: Create an evaluation matrix consisting of m alternatives and n criteria, with the intersection of each alternative and criteria given as $x_{ij}$, we therefore have a matrix $(x_{ij})_{m\times n}$
\STATE \textbf{Step 2}: Normalize the matrix $(x_{ij})_{m\times n}$
\STATE \textbf{Step 3}: Calculate the weighted normalized decision matrix 
\STATE \textbf{Step 4}: Determine the worst alternative and the best alternative
\STATE \textbf{Step 5}: Calculate the L2 distance between the target alternative $i$ and the worst condition $A_w$
\STATE \textbf{Step 6}: Calculate the similarity to the worst condition 
\STATE \textbf{Step 7}: Rank the alternatives according to $s_{iw}$
\end{algorithmic}
\caption{TOPSIS Method}
\label{alg:topsis}
\end{algorithm}

\end{document}